\documentclass{svproc}
\usepackage{url}

\usepackage{marvosym}
\usepackage{amsmath}
\usepackage{bm}
\usepackage{graphicx}
\usepackage{amssymb}
\usepackage{booktabs}
\usepackage{multirow}
\usepackage{placeins}
\usepackage{siunitx}
\begin{document}
\mainmatter{}     % start of a contribution
\title{The Setting of IMU Parameters in Kalman Filtering-based Information Fusion}
\author{Qiang Hu \and Yanhua Zou \and Shuaiyi Huo \and Haibo Ge \textsuperscript{(\Letter)} \and Wei Ouyang\textsuperscript{(\Letter)}}


%
\institute{Tongji University, Shanghai, China, 200092\\
\texttt{Corresponding Author:Haibo Ge, Wei Ouyang}\\
\email{haibo\_ge@tongji.edu.cn, ywoulife@tongji.edu.cn}
}

\maketitle              % typeset the title of the contribution

\begin{abstract}
The setting or tuning of specifications for the inertial measurement unit (IMU) is tricky in sensor fusion. The underneath conundrum is caused by the fact that the working condition of IMU is more complex than the stationary calibration scenario. Since the noises and biases instabilities calibrated under static condition cannot accommodate other cases, the effective tuning of IMU parameters largely hinges on the experience or profound understanding of the system. In the current work, the setting method of IMU parameters based on Allan variance calibration is delved into within the Kalman filtering framework. Specifically, the relationship between the power sepctral density and Allan variance is leveraged in formulating the process uncertainty in continuous-time filtering. Three typical IMU-based sensor fusion systems, including INS/GNSS integration, LiDAR-inertial odometry, and visual-inertial odometry are considered to show the feasibility and effectiveness of this parameter setting process.

% We would like to encourage you to list your keywords within
% the abstract section using the \keywords{...} command.
\keywords{Inertial Measurement Unit, Kalman Filtering, Information Fusion, Noise, Bias Instability}
\end{abstract}
\section{Introduction}
%     Problem statement

The Inertial Measurement Unit (IMU) is known for its independence on exterior measurements, providing high-frequency motion information which is pivotal in integrating other sensors' measurements. The IMU is usually integrated with Global Navigation Satellite System (GNSS), Light Detection And Ranging sensor (LiDAR), camera and Doppler Velocity Log (DVL), etc. The grades of IMU range from tactical, navigation, industrial and consumer grade primarily according to the precision of gyroscope. Two dominated specifications of the gyroscope and accelerometer are the noise and bias instability, perplexing novices and practitioners when designing a new navigation system. The main predicament is how to set the corresponding gyroscope and accelerometer parameters in Kalman filtering-based or optimization-based navigation algorithms.

Noises and bias instabilities describe the uncertainty of process model, which take effect in balancing the contributions of propagation and measurement update. Inappropriate settings of noise and bias instability parameters lead to sub-optimal information fusion. In inertial-based information fusion, the measurements from GNSS and LiDAR are output in lower frequency than IMU and the uncertainty of process model is propagated with the same frequency of IMU. However, the primitive kinematic model is continuous and nonlinear, and the noises are modeled in the continuous-time domain, i.e., the power spectral density (PSD). The gap between the continuous- and discrete-time descriptions of IMU uncertainties is abstruse. The most widely used tool to quantify the specifications of IMU is the Allan variance, which is computed with the static IMU data. Since proposed, researchers have investigated different approaches to obtaining detailed specifications of IMU through fitting the Allan variance plot. For instance, we know that the slope $-1/2$ corresponds to the noise random walk ($N$), the slope 0 corresponds to the bias instability ($B$) and the slope $1/2$ corresponds to the rate random walk ($K$), in which, $N$ and $B$ are important specifications in inertial-based sensor fusion. In the current work, the relationship between the PSD and calibrated specifications is formulated based on the tutorial \cite{farrell}.

In contrast with setting the IMU parameters based on the calibration results, researchers also investigated the estimation approach for the parameters related with process noises.  Covariance matching methods were initially used to estimate the noise covariance matrices by the innovation covariance matrix in INS/DVL~\cite{Gao} and INS/GNSS~\cite{HUG} integrated systems. Huang proposed the adaptive Kalman filtering method based on the nonadjacent state transition model for estimating the small-magnitude and inaccurate process noise covariance matrix, showing improved estimation accuray of noise parameter; however, the estimation of bias instability is left as future work~\cite{zhu2025adaptive}. A reduced-order multiple-model (MM) estimator for process and measurement noise identification was proposed in~\cite{Khalife}, which improved the computational efficiency in contrast with the standard MM method. Nevertheless, the estimation performance directly depends on the reliability of designed parameter set for noises. Reinforcement learning methods have also been proposed for inertial-based integrated navigation, such as the reinforcement learning adaptive Kalman filter~\cite{GaoRLEKF} and the dynamic grid-based Q-learning EKF~\cite{Dai_qlekf}. Deep neural networks were designed in~\cite{Brossard} to dynamically adapt the noise parameters of the Kalman filter in IMU-only dead-reckoning. In general, the estimated or calibrated noise characteristics of IMU tend to be reliable under different working conditions, whereas estimating the bias instability is more tricky. 

Above works estimate the constant or dynamic noise parameters of IMU, but for industrial-grade or consumer-grade inertial sensors the bias instability is less observable and hard to be estimated online. In this work, we consider the setting and tuning of IMU parameters as per the specifications provided by manufacturers or calibrated by Allan variance. The main contribution includes:
\begin{enumerate}
    \item the continuous-time and discrete-time noise matrices are revisited, and the setting and tuning of process covariance matrix according to the Allan variance calibration is proposed;
    \item the proposed method is tested in three representative systems: INS/GNSS integration, LiDAR-inertial odometry, and visual-inertial odometry.
\end{enumerate}
The subsequent Section 2 revisits the relationship between continuous-time and discrete-time process models for error-state Kalman filter. The setting and tuning method for IMU parameters in process noise covariance matrix is given in Section 3. Section 4 performs experimental verifications based on INS/GNSS, LiDAR-inertial, and visual-inertial systems. Finally, Section 5 concludes this paper.
%==============================================================
\section{Continuous- to Discrete-Time Filtering Models}

In this section, we will reexamine the approaches for discretizing the continuous-time process model for inertial-based navigation problem. As for a general nonlinear continuous-time system model
\begin{equation}
\label{system}
{\mathbf{\dot x}} = {\bf{f}}\left( {{\bf{x}},{\bf{t}},{\bf{u}}} \right)
\end{equation}
in which $\bf{u}$ denotes the inputs from gyroscopes and accelerometers in inertial-based navigation, $\bf{f}$ is the nonlinear kinematic model. 

Considering the fact that the attitude of a rigid body is parameterized by the rotational matrix or quaternion, the state $\bf{x}$ typically includes $\{\bf{R},\bf{v},\bf{p},\bf{b_g},\bf{b_a}\}$, in which $\bf{R}\in\it{SO}\textrm{(3)}$, $\bf{v},\bf{p},\bf{b_g},\bf{b_a}\in{\mathbb{R}^\textrm{3}}$.
Performing the linearization on Eq. (\ref{system}) leads to the error-state differential equation, which describes the local characteristics of the navigation system.
\begin{equation}
\label{error_diff}
    \delta {\mathbf{\dot x}} = {\mathbf{f}}\left( {\delta {\mathbf{x}},t,\delta {\mathbf{u}}} \right)
\end{equation}
where the error state includes the composition of errors derived from the general minus operation $\delta {\mathbf{x}} = {\mathbf{x}} - {\mathbf{\hat x}}$. Specifically, the Lie algebra of group error and the errors in vector space are concatenated as the error-state vector.

Note that the input $\mathbf{u}$ contains the measured angular velocity and the specific force by the IMU
\begin{equation}
\begin{array}{rcl}
{\bf{\tilde \omega }}_{ib}^b = {\bf{\omega }}_{ib}^b + {{\bf{b}}_g} + {{\bf{w}}_g},
{\bf{\tilde f}}_{ib}^b = {\bf{f}}_{ib}^b + {{\bf{b}}_a} + {{\bf{w}}_a}.
\end{array}
\end{equation}
where ${{\bf{w}}_g},{{\bf{w}}_a}$ are the noises of gyroscope and accelerometer. 

The biases of IMU are modeled as the first-order Gauss-Markov process
\begin{equation}
\label{gmmodel}
\begin{array}{rcl}
{{{\bf{\dot b}}}_g} =  - \frac{1}{{{T_{gb}}}}{{\bf{b}}_g} + {{\bf{w}}_{{g_b}}}, 
{{{\bf{\dot b}}}_a} =  - \frac{1}{{{T_{ab}}}}{{\bf{b}}_a} + {{\bf{w}}_{{a_b}}}.
\end{array}
\end{equation}
in which $T_{gb},T_{ab}$ are the correlation time.

The continuous-time white noises ${\bf{\delta u}} = {\left[ {{{\bf{w}}_g}^T,{{\bf{w}}_a}^T,{{\bf{w}}_{{b_g}}}^T,{{\bf{w}}_{{b_a}}}^T} \right]^T}$ are described by the power spectral density
\begin{equation}
E\left[ {{\bf{\delta u}}(t){\bf{\delta }}{{\bf{u}}^{\bf{T}}}(\tau )} \right] = {\bf{S}}\delta (t - \tau )
\end{equation}
in which ${\bf{S}}$ is the power spectral density.

The error-state model (\ref{error_diff}) can be further linearized with respect to the error-state and process noises as below
\begin{equation}
\label{continuous}
\begin{aligned}
\delta \dot{\mathbf{x}}(t) 
&= \frac{\partial \mathbf{f}}{\partial \delta \mathbf{x}} \delta \mathbf{x}(t) + \frac{\partial \mathbf{f}}{\partial \delta \mathbf{u}} \mathbf{\delta u}(t)
= \mathbf{F}(t) \delta \mathbf{x}(t) + \mathbf{G}(t) \mathbf{\delta u}(t)
\end{aligned}
\end{equation}
where the matrix ${\bf{G}}\left( t \right)$ is the input matrix for additive process noises of IMU.

Since discrete-time covariance propagation is mostly considered in indirect Kalman Filtering, the continuous model (\ref{continuous}) should be discretized to
\begin{equation}
\label{onetype}
   \delta {{\bf{x}}_k} = {{\bf{\Phi }}_{k|k - 1}}\delta {{\bf{x}}_{k - 1}} + {{\bf{W}}_{k - 1}}
\end{equation}
where ${{\bf{\Phi }}_{k|k - 1}} = \exp \left( {\int_{{t_{k - 1}}}^{{t_k}} {{\bf{F}}\left( t \right)} dt} \right)\delta {{\bf{x}}_{k - 1}},{{\bf{W}}_{k - 1}} = \int_{{t_{k - 1}}}^{{t_k}} {{{\bf{\Phi }}_{k|t}}{\bf{G}}\left( t \right){\bf{\delta u}}\left( t \right)} dt$. 

Therefore, the discrete-time process noise covariance matrix can be computed as \cite{Brown}
\begin{equation}
\label{Q_expectation}
\begin{aligned}
{{\bf{Q}}_{k-1}} &= E\left\{ {{{\bf{W}}_{k - 1}}{\bf{W}}_{k - 1}^T} \right\}\\
& = E\left\{ {\int_{{t_{k - 1}}}^{{t_k}} {{{\bf{\Phi }}_{k|t}}} {\bf{G}}(t){\bf{\delta u}}(t){\bf{d}}t{{\left[ {\int_{{t_{k - 1}}}^{{t_k}} {{{\bf{\Phi }}_{k|\tau }}} {\bf{G}}(\tau ){\bf{\delta u}}(\tau ){\bf{d}}\tau } \right]}^T}} \right\}\\
& \approx \frac{1}{2}\left[ {{{\bf{\Phi }}_{k/k - 1}}{\bf{G}}\left( {{t_{k - 1}}} \right){\bf{S}}{{\bf{G}}^{\it{T}}}\left( {{t_{k - 1}}} \right){\bf{\Phi }}_{k/k - 1}^{\it{T}} + {\bf{G}}\left( {{t_k}} \right){\bm{S}}{{\bf{G}}^{\it{T}}}\left( {{t_k}} \right)} \right]\Delta t.
\end{aligned}   
\end{equation}

In practice, we can regard ${\bf{G}}\left( {{t_{k - 1}}} \right) = {\bf{G}}\left( {{t_k}} \right)$ for two adjacent states. If we further assume the state transition matrix ${{\bf{\Phi }}_{k/k - 1}} \approx {{\bf{I}}_{15}}$ for small interval $\Delta t$, then 
\begin{equation}
\label{continuous_q}
    {{\bf{Q}}_{k-1}} \approx {\bf{G}}\left( {{t_{k-1}}} \right){\bf{S}}\Delta t{{\bf{G}}^T}\left( {{t_{k-1}}} \right).  
\end{equation}

The above derivations are named as {\it{linearization-and-discretization}} formulation. One may first discretize the differential equation (\ref{continuous}) and linearize it with respect to error states. In such case, the derivations involve with the discrete noises. For small $\Delta (t)$, the differential equation (\ref{error_diff}) can be discretized in a sophisticated way or directly discretized by using the Euler integration
\begin{equation}
\label{another}
\begin{aligned}
\delta {{\bf{x}}_k} & \approx \left( {{\bf{I}} + {\bf{F}}\left( {{t_{k - 1}}} \right)\Delta t} \right)\delta {{\bf{x}}_{k - 1}} + {\bf{G}}\left( {{t_{k - 1}}} \right)\Delta t{\bf{w}}\left( {{t_{k - 1}}} \right)\\
& \approx {{\bf{\Phi }}_{k/k - 1}}\delta {{\bf{x}}_{k - 1}} + {{\bm{\eta }}_{k - 1}}
\end{aligned}
\end{equation}
where ${{\bf{F}}\left( {{t_{k - 1}}} \right)}$,${\bf{G}}\left( {{t_{k - 1}}}\right)$ denote the discrete-time Jacobian matrices, and ${\bf{w}}\left( {{t_{k - 1}}}\right)$ is the  white noise at $t_{k}$ due to the white-noise input $\delta {\bf{u}} \left( t\right)$ during the $\;\left( {{t_{k - 1}},{t_k}} \right)$ interval \cite{Brown}.

In such case, the discrete process noise covariance matrix is accordingly computed as
\begin{equation}
\label{Q2}
\begin{aligned}
{{\bf{Q}}_{k - 1}} & = E\left\{ {{{\bf{\eta }}_{k - 1}}{\bf{\eta }}_{k - 1}^T} \right\}\\
& = {\bf{G}}\left( {{t_{k - 1}}} \right)E\left\{ {{\bf{w}}\left( {{t_{k - 1}}} \right){\bf{w}}{{\left( {{t_{k - 1}}} \right)}^T}} \right\}{\bf{G}}{\left( {{t_{k - 1}}} \right)^T}\Delta {t^2}.
\end{aligned}
\end{equation}
and such {\it{discretization-and-linearization}} formulation  is adopted in the open-source code of FASTLIO2 \cite{fastlio2}.

Note that (\ref{another}) should be equivalent to (\ref{onetype}) and according to (\ref{continuous_q}) and (\ref{Q2}) we have
\begin{equation}
\label{Qk_with_S}
E\left\{ {{\bf{w}}\left( {{t_{k - 1}}} \right){\bf{w}}{{\left( {{t_{k - 1}}} \right)}^T}} \right\} = \frac{{\bf{S}}}{{\Delta t}}
\end{equation}
which means that the discrete noise covariance matrix is related with the power spectral density by the sampling interval. 

The setting or tuning the elements in power spectral density matrix is pivotal for the estimation performance of Kalman filtering-based information fusion algorithms. In the next section, the parameter setting method is introduced according to the widely used Allan variance calibration.

% ===================setting method =====================
\section{The Parameter Setting by Allan Variance}

The focused specifications obtained by the Allan variance are the noise random walk and bias instability, which will be used to set the power spectral density matrix. First, we revisit the computation of Allan variance based on the power spectral density. 
\begin{equation}
\label{stosigma}
    \sigma _u^2(\tau ) = 4\int_0^\infty  {{S_u}} (f)\frac{{{{\sin }^4}(\pi f\tau )}}{{{{(\pi f\tau )}^2}}}df
\end{equation}
where ${S_u}(f) = {\left. {{S_u}(s)} \right|_{s = j2\pi f}}$ is the power spectral density for the signal $u$, $s \in \mathbb{C}$ is the Laplace variable, $f=1/T$ is the frequency in Hertz, $\tau=nT$ is the length of data used in computing the Allan variance.

The power spectral density can be modeled as a  series of frequency $f$
\begin{equation}
    {S_u}(f) =  \cdots  + {N^2} + \frac{{{B^2}}}{{2\pi f}} + \frac{{{K^2}}}{{{{(2\pi f)}^2}}} +  \cdots 
\end{equation}
in which $N,B,K$ are the coefficients corresponding to three components of the stochastic signal and Allan variance
\begin{equation*}
\begin{aligned}
u(t) &=  \cdots  + {z_N}(t) + {z_B}(t) + {z_K}(t) +  \cdots  \\
\sigma _u^2(\tau ) &=  \cdots  + \sigma _{{z_N}}^2(\tau ) + \sigma _{{z_B}}^2(\tau ) + \sigma _{{z_K}}^2(\tau ) +  \cdots 
\end{aligned}
\end{equation*}

As for the noises, substitute ${S_u}\left( f \right) = {N^2}$ into (\ref{stosigma}) and we can obtain $\sigma _{{z_N}}^2 = \frac{{{N^2}}}{\tau }$. Similarly, the Allan variance of bias instability can be computed by substituting ${S_u}\left( f \right) = \frac{{{B^2}}}{{2\pi f}}$ into (\ref{stosigma}) to get
\begin{equation}
    \sigma _{{z_B}}^2 = \frac{{2{B^2}}}{\pi }\int_0^\infty  {\frac{{{{\sin }^4}\left( {\pi f\tau } \right)}}{{{{\left( {\pi f\tau } \right)}^2}}}} d\left( {\pi f\tau } \right)
\end{equation}
and according to the integral rule $\int_0^\infty  {\frac{{{{\sin }^4}ax}}{{{x^3}}}} dx = {a^2}\ln 2$ for $a=1$, we further have
\begin{equation}
\label{sigma_zb}
\begin{aligned}
     \sigma _{{z_B}}^2 & = \frac{{2{B^2}\ln (2)}}{\pi } \hfill  \approx {(0.664B)}^2 \hfill \\  
\end{aligned}
\end{equation}

Note that the variance for bias instability is independent of $\tau$, and therefore, the value of $\sigma _{{z_B}}^2$ corresponds to the plot of Allan variance with zero slope.

Considering the first-order Gauss-Markov models in (\ref{gmmodel}), the general continuous bias process is
\begin{equation}
{\dot z_B}(t) =  - {\mu _B}{z_B}(t) + {n_B}(t),{\mu _B} = \frac{1}{{{T_B}}}
\end{equation}
where $n_B$ is the driving noise with power spectral density $S_B$.

The power spectral density of $z_B$ is 
\begin{equation}
{S_{{z_B}}}(\omega ) = \frac{{{S_B}}}{{{\omega ^2} + \mu _B^2}}, \omega =2\pi f
\end{equation}

 Substitute ${S_{{z_B}}}(\omega )$ into (\ref{stosigma}), we have \cite{standard}
\begin{equation}
\sigma _{{z_B}}^2(\tau ) = \frac{{{S_B}T_B^2}}{\tau }\left[ {1 - \frac{{{T_B}}}{{2\tau }}\left( {3 - 4{e^{ - \frac{\tau }{{{T_B}}}}} + {e^{ - \frac{{2\tau }}{{{T_B}}}}}} \right)} \right].
\end{equation}
which is related with the ratio $\lambda  = {\tau  /{{T_B}}}$, and then, we can transform it as $\sigma _{{z_B}}^2(\tau ) = {S_B}{T_B}F\left( \lambda  \right),F\left( \lambda  \right) = \frac{1}{\lambda } - \frac{1}{{2{\lambda ^2}}}\left( {3 - 4{e^{ - \lambda }} + {e^{ - 2\lambda }}} \right)$.

\begin{figure}[ht!]
\vspace{-1em}  
\centering
\includegraphics[width=0.65\textwidth]{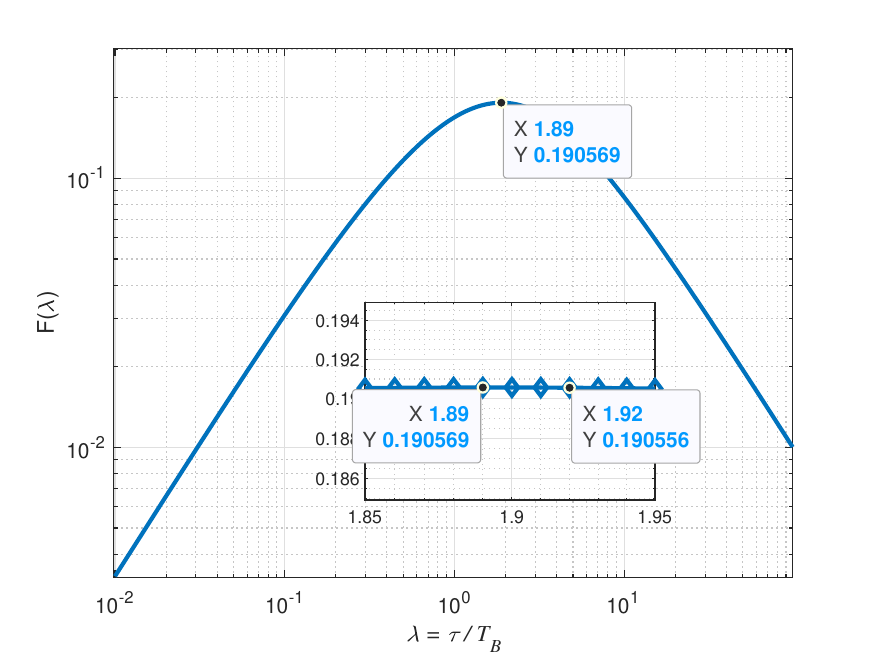}
\caption{The variation of $\sigma _{{z_B}}^2(\tau )$ with the ratio $\lambda$}
\label{figure_sb}
\end{figure}

Here, we can plot $F\left( \lambda  \right)$ in Fig. \ref{figure_sb} and check $\lambda$ corresponding to the zero slope. When $\lambda \approx 1.89, F(1.89) \approx 0.19$, and therefore
\begin{equation}
\sigma _{{z_B}}^2\left( {1.89{T_B}} \right) = {\left( {0.4365\sqrt {{S_B}{T_B}} } \right)^2}
\end{equation}
and according to (\ref{sigma_zb}), we can get the power spectral density 
\begin{equation}
\label{new_S}
\begin{aligned}
  {S_B} & \approx \frac{{2{B^2}\ln (2)}}{{\pi {{\left( {0.4365} \right)}^2}{T_B}}} \hfill \approx \frac{{5.25\sigma _{{z_B}}^2}}{{{T_B}}} \hfill \\ 
\end{aligned}
\end{equation}
where $\sigma_{z_B}$ can be obtained from the Allan variance plot.

\begin{figure}[ht!]
\vspace{-0.5cm}
\centering
\includegraphics[width=0.99\textwidth]{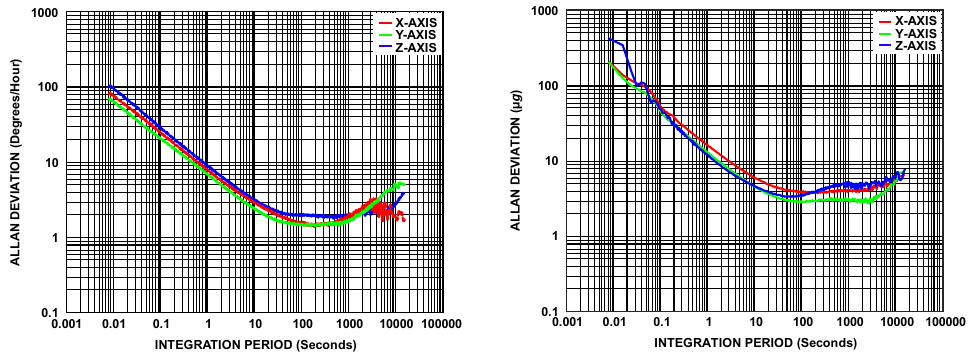}
\caption{The Allan deviation plot of ADIS16465-1 (see footnote 1).}
\label{adi_allan}
\vspace{-0.6cm}
\end{figure}

Take the ADIS16465-1 IMU of Analog devices as an example\footnote{https://www.analog.com/media/en/technical-documentation/data-sheets/adis16465.pdf}, whose correlation time of each axis can be extracted from the Allan deviation plot shown in Fig. \ref{adi_allan}. For the $X,Y$ axes of gyroscopes, $\tau = 200$s and the correlation time ${T_B} = 105.8{\text{s}}$, ${\sigma _{bg}} = {{2^ \circ }}/h$, $B=3^\circ/h$, ${S_{Bg}} = {\text{1.53e-08}}(^\circ/\text{s})^2/\text{s}$. For the $Z$ axis, the flat region ranges from 400s to
1000s. Here we use $\tau \approx 1000$s and the correlation time ${T_B} = 529.1{\text{s}}$, ${S_{Bg}} = {\text{3.06e-09}}(^\circ/\text{s})^2/\text{s}$. For accelerometers, the flat region of $X,Y$ axes ranges from $\tau=200$s to $\tau=3000$s. If we take $\tau=200$s, then ${T_B} = 105.8{\text{s}}$, ${\sigma _{ba}} = {{3.6\mu g}}$, $B=5.4\mu g$, ${S_{Ba}} = {\text{6.431e-11}}(m/\text{s}^2)^2/\text{s}$. Similarly, for $Z$ axis $\tau = 80$s, ${T_B} = 42.3{\text{s}}$, ${S_{Ba}} = {\text{1.6e-10}}(m/\text{s}^2)^2/\text{s}$.

It can be seen that the uncertainty on the correlation time $T_B$ of each axis will unavoidably affect the reliability of $S_B$, and the resultant power spectral density would differ a lot from the most appropriate value. Hence, specific tuning on $S_B$ is necessary for make full use of the IMU data. Since the calibrated Allan variance and correlation time are non-isotropic for gyroscopes and accelerometers, the power spectral density matrix is formulated as
\begin{equation}
\begin{aligned}
  {\mathbf{S}} & = \left[ {\begin{array}{*{20}{c}}
  {{{\mathbf{N}}_g}}&0&0&0 \\ 
  0&{{{\mathbf{N}}_a}}&0&0 \\ 
  0&0&{{{\mathbf{S}}_{Bg}}}&0 \\ 
  0&0&0&{{{\mathbf{S}}_{Ba}}} 
\end{array}} \right], \hfill \\
  {{\mathbf{N}}_g} & = diag\left[ {\begin{array}{*{20}{c}}
  {{N_{gx}}}&{{N_{gy}}}&{{N_{gz}}} 
\end{array}} \right], 
  {{\mathbf{N}}_a} = diag\left[ {\begin{array}{*{20}{c}}
  {{N_{ax}}}&{{N_{ay}}}&{{N_{az}}} 
\end{array}} \right], \hfill \\
  {{\mathbf{S}}_{Bg}} & = 5.25 \cdot diag\left[ {\begin{array}{*{20}{c}}
  {\frac{{\sigma _{bgx}^2}}{{{T_{Bgx}}}}}&{\frac{{\sigma _{bgy}^2}}{{{T_{Bgy}}}}}&{\frac{{\sigma _{bgz}^2}}{{{T_{Bgz}}}}} 
\end{array}} \right], \hfill 
  {{\mathbf{S}}_{Ba}}   = 5.25 \cdot diag\left[ {\begin{array}{*{20}{c}}
  {\frac{{\sigma _{bax}^2}}{{{T_{Bax}}}}}&{\frac{{\sigma _{bay}^2}}{{{T_{Bay}}}}}&{\frac{{\sigma _{baz}^2}}{{{T_{Baz}}}}} 
\end{array}} \right]. \hfill \\ 
\end{aligned} 
\end{equation}

There following tuning strategy for the power spectral density matrix is adopted.

\begin{equation}
\begin{gathered}
  \left[ {\begin{array}{*{20}{c}}
  {{\sigma _{bgx}}}&{{\sigma _{bgy}}}&{{\sigma _{bgz}}} 
\end{array}} \right] \triangleq {k_{bg}}\left[ {\begin{array}{*{20}{c}}
  {{\sigma _{bgx}}}&{{\sigma _{bgy}}}&{{\sigma _{bgz}}} 
\end{array}} \right], \hfill \\
  \left[ {\begin{array}{*{20}{c}}
  {{\sigma _{bax}}}&{{\sigma _{bay}}}&{{\sigma _{baz}}} 
\end{array}} \right] \triangleq {k_{ba}}\left[ {\begin{array}{*{20}{c}}
  {{\sigma _{bax}}}&{{\sigma _{bay}}}&{{\sigma _{baz}}} 
\end{array}} \right]. \hfill \\ 
\end{gathered} 
\end{equation}

Note that the  FASTLIO2 \footnote{\url{https://github.com/hku-mars/FAST_LIO}} set the discrete-time covariance of noise random walks as $Q_g = 0.1(\text{deg})^2, Q_a = 0.1(\text{m/s})^2$ bias noises as $Q_{bg}=0.0001 (\text{deg}/\text{s})^2$, $Q_{ba}=0.0001 (\text{m}/\text{s}^2)^2$.

% ========================================
\section{Experiments}
In this section, the tuning approach will be verified in INS/GNSS, LiDAR-inertial, and visual-inertial systems. As for Kalman filtering-based INS/GNSS fusion algorithm, the open-source code KFGINS \footnote{\url{https://github.com/i2Nav-WHU/KF-GINS}} and its companion data are used. The calibrated specifications of the ADIS16465 IMU have been given before. Although the specific type of IMU \{16465-1, 16465-2\} is unknown, the specifications of accelerometers are identical for them and the specifications of gyroscopes are also similar. Note that the built-in Kalman filter defined the power spectral density matrix as ${\mathbf{q}} = diag\left[ {N_g^2{{\mathbf{I}}_3},N_a^2{{\mathbf{I}}_3},2\sigma _{gb}^2{{\mathbf{I}}_3}/{T_{Bg}},2\sigma _{ab}^2{{\mathbf{I}}_3}/{T_{Ba}}} \right]$, and the parameters were set as $N_g=0.1\text{deg}/\sqrt{\text{h}}$, $N_a=0.1\text{m/s}/\sqrt{\text{h}}$, $\sigma_{gb} = 50\text{deg}/\text{h}$,$\sigma_{ab}=50\text{mGal}$ in KFGINS. According to the official manual of this IMU, $N_g=0.15\text{deg}/\sqrt{\text{h}}$, $N_a=0.012\text{m/s}/\sqrt{\text{h}}$, $\sigma_{gb} = 2\text{deg}/\text{h}$,$\sigma_{ab}=3.6\mu \text{g}$. Here, the parameters of noise random walks are set according to calibrated specifications, but the bias instabilities are enlarged to $\sigma_{gb} = k_{gb}*2\text{deg}/\text{h}$,$\sigma_{ab}=k_{ab}*3.6\mu \text{g}$.

The results are compared in Table \ref{INSGNSS}, which indicate that this system prefers the standard setting of calibrated parameters. As for smaller parameters $ k_{gb}=k_{ab}=0.5$, it also achieves satisfactory positioning accuracy. It might be owing to the fact that suppressing the divergence of biases during the GNSS outages is beneficial to the positioning accuracy.

\begin{table}[ht!]
\caption{RMSEs of states for built-in and tuned parameters KF-GINS}
\vspace{-1.5\baselineskip}  
\label{INSGNSS}
\begin{center}
\begin{tabular}{c c @{\hspace{0.5cm}} c}
\hline
Tuning & Attitude (deg, [Roll Pitch Yaw]) & Position (m, [N, U, E]) \\
\hline
Built-in & [0.035	0.053	0.385] & [0.387	0.711	0.383]  \\
\hline
$ k_{gb}=k_{ab}=0.5$ & \textbf{[0.027	0.052	0.367]} & \textbf{[0.236 0.463	0.409]}  \\
\hline
$ k_{gb}=k_{ab}=1$  & \textbf{[0.027	0.052	0.367]} & [0.246	0.484	0.409]  \\
\hline
$ k_{gb}=k_{ab}=3$  & [0.028	0.052	0.380] & [0.318	0.569	0.409] \\
\hline
$ k_{gb}=k_{ab}=5$  & [0.031	0.053	0.402] & [0.362	0.653	0.408]  \\
\hline
\end{tabular}
\end{center}
\vspace{-2\baselineskip}
\end{table}

As for the inertial-LiDAR information fusion, the open-source FASTLIO2 is used to verify the effectiveness of parameter tuning method. Considering the fact that LiDAR provides strong corrections to INS in structured man-made scenarios, we performed the tests under degenerated scenarios where the potential of IMU can be leveraged well to improve the positioning accuracy. The degenerated dataset \cite{degenerate} used the Xsens MTi-600 IMU in date collection, in which the sequence 4 is the most challenging. This type of IMU is also calibrated by the authors to obtain the specifications. The power spectral density of noise random walks and bias instabilities are $N_g=0.42\text{deg}/\sqrt{\text{h}}$, $N_a=60\mu \text{g}/\sqrt{\text{Hz}}$, $T_{Bg}=31.7\text{s}$, $\sigma_{gb} = 8\text{deg}/\text{h}$, $T_{Ba\{x,y\}}=63.5\text{s}$, $T_{Ba\{z\}}=31.7\text{s}$, $\sigma_{ab}=10\mu \text{g}$, and different tuning parameters $k_{gb}, k_{ab}$ are examined. Note that the algorithm is run for 10 times for each setting of parameters to obviate the randomness. The maximum, minimum and average of RMSEs are given in Table \ref{FASTLIO2-1}. The tuning approach shows improved positioning accuracy in Seq1, Seq4 and Seq7.

\begin{table}[ht!]
\caption{RMSEs of position  errors for FASTLIO2 and tuned parameters}
\vspace{-1.5\baselineskip}  
\label{FASTLIO2-1}
\begin{center}
\begin{tabular}{c c @{\hspace{0.5cm}} c}
\hline
Tuning & Seq1([Max Min Average]m) & Seq2 ([Max Min Average]m) \\
\hline
FASTLIO2 & [2.166 1.535 1.845] & [\textbf{0.511} \textbf{0.470} \textbf{0.490}]  \\
\hline
$ k_{gb}=k_{ab}=10$ & [\textbf{1.587} 1.409 1.490] & [0.600 0.542 0.572]  \\
\hline
$ k_{gb}=k_{ab}=25$  & [1.617 \textbf{1.360} \textbf{1.458}] & [0.689 0.575 0.644]  \\
\hline
$ k_{gb}=k_{ab}=50$  & [1.691 1.456 1.553] & [0.693 0.633 0.678] \\
\hline
{} & Seq4([Max Min Average]m) & Seq7 ([Max Min Average]m) \\
\hline
FASTLIO2 & [22.643 6.901 11.709] & [0.211 \textbf{0.144} 0.188]  \\
\hline
$ k_{gb}=k_{ab}=10$ & [diverge	diverge	diverge] & [diverge	diverge	diverge]  \\
\hline
$ k_{gb}=k_{ab}=25$  & [29.459 3.439 18.407] & [\textbf{0.169} 0.147 \textbf{0.163}]  \\
\hline
$ k_{gb}=k_{ab}=50$  & [\textbf{6.618} \textbf{3.212} \textbf{5.582}] & [0.221 0.156 0.178] \\
\hline
{} & Seq8([Max Min Average]m)  \\
\hline
FASTLIO2 & [\textbf{0.820 0.662 0.729}]   \\
\hline
$ k_{gb}=k_{ab}=10$ & [1.051 0.807 0.987]  \\
\hline
$ k_{gb}=k_{ab}=25$  & [0.996 0.913 0.952]   \\
\hline
$ k_{gb}=k_{ab}=50$  & [0.915 0.751 0.8414] \\
\hline
\end{tabular}
\end{center}
\vspace{-2\baselineskip} 
\end{table}

We also performed more complex tuning by first scrutinizing the influence of different $k_{ab},k_{gb}$ on the position accuracy through multiple trials, which conduct brute-force testing the combinations of $k_{ab}=\{10, 15,20,25,30,40,45,50\}, k_{gb}=\{10, 15,20,25,30,40,45,50\}$. We finally select the most suitable companion $k_{ab}=40,k_{gb}=20$ and test this setting under these sequences. Ten runs are performed on each sequence and averaged results are given in the following Table \ref{FASTLIO2-3}.

\begin{table}[ht!]
\caption{RMSEs of position  errors for FASTLIO2 with tuned parameters }
\vspace{-1.5\baselineskip}
\label{FASTLIO2-3}
\begin{center}
\begin{tabular}{c c @{\hspace{0.5cm}} c}
\hline
Tuning & Seq1([Max Min Average]m) & Seq2 ([Max Min Average]m) \\
\hline
FASTLIO2 & [2.166 1.535 1.845] & [\textbf{0.511 0.470 0.490}] \\
\hline
$ k_{gb}=20, k_{ab}=40$ & [\textbf{1.662 1.506 1.580}] & [0.689 0.563 0.637] \\
\hline
{} & Seq4([Max Min Average]m) & Seq7 ([Max Min Average]m) \\
\hline
FASTLIO2 & [22.643 6.901 11.709] & [0.211 \textbf{0.144} 0.188] \\
\hline
$ k_{gb}=20, k_{ab}=40$ & [\textbf{4.264 2.528 3.261}] & [\textbf{0.189} 0.173 \textbf{0.183}] \\
\hline
{} & Seq8([Max Min Average]m) \\
\hline
FASTLIO2 & [\textbf{0.820 0.662 0.729}]    \\
\hline
$ k_{gb}=20, k_{ab}=40$ & [1.082 0.861 0.935]  \\
\hline
\end{tabular}
\end{center}
\vspace{-2\baselineskip}  
\end{table}

\begingroup
\emergencystretch=1em
Note that the FASTLIO2 sets the discrete covariance of bias instabilities as $cov_{bg}=0.0001 (\rm{deg/h})^2, cov_{ba}=0.0001 (\rm{m/s^2})^2$. Based on the discrete-continuous relationship in (\ref{Qk_with_S}) and (\ref{new_S}), the computed coefficients for Allan variances are approximately $k_{gb} \approx 32.5,k_{ab}\approx 12.5$ for 400Hz IMU frequency, which show sharp contrast with the tuned parameters in this work. As for the most challenging sequence 4, the standard FASTLIO2 performs mush worse than the tuned parameters.
\par
\endgroup
\FloatBarrier

\setlength{\skip\footins}{1.5\baselineskip}
As for visual-inertial navigation, the open-source OpenVINS \footnote{\url{https://github.com/rpng/open_vins}} \cite{openvins} is evaluated on the \texttt{building02} sequence of the i2Nav-Robot dataset \cite{i2navrobot}, which uses an ADIS16465 IMU and an AVT Mako-G234 stereo camera. The IMU power spectral densities are set to the same values as in the preceding KFGINS experiment. Since the IMU plays a more important role in visual-inertial odometry, this experiment simultaneously adjusts the coefficients of the IMU measurement noises and bias instabilities. Specifically, $k_a$ and $k_g$ denote the scale coefficients applied to the accelerometer and gyroscope measurement noises, respectively, whereas $k_{ab}$ and $k_{gb}$ scale the corresponding bias-instability parameters.

The four coefficients are tuned using a sequential search strategy. First, $k_g$, $k_{ab}$, and $k_{gb}$ are fixed at 1 while searching for the optimal $k_a$. The selected value of $k_a$ is then retained while $k_g$ is searched, followed successively by searches for $k_{ab}$ and $k_{gb}$, with the previously selected coefficients kept fixed at each step.

\begin{table}[ht!]
\caption{Position RMSEs of OpenVINS with tuned IMU parameters}
\vspace{-1.5\baselineskip}
\label{OpenVINS}
\begin{center}
\begin{tabular}{c c c c c c}
\hline
Dataset & $k_a$ & $k_g$ & $k_{ab}$ & $k_{gb}$ & RMSE (m) \\
\hline
\multirow{5}{*}{building02} & OpenVINS & OpenVINS & OpenVINS & OpenVINS & 1.483 \\
\cline{2-6}
& 40 & 1 & 1 & 1 & 1.443 \\
\cline{2-6}
& 40 & 90 & 1 & 1 & 0.874 \\
\cline{2-6}
& 40 & 90 & 10 & 1 & 0.872 \\
\cline{2-6}
& 40 & 90 & 10 & 10 & \textbf{0.870} \\
\hline
\end{tabular}
\end{center}
\vspace{-2\baselineskip}
\end{table}

As shown in Table \ref{OpenVINS}, the sequential search yields the final coefficient combination $(k_a,k_g,k_{ab},k_{gb})=(40,90,10,10)$. Compared with the default OpenVINS parameters, this combination reduces the position RMSE from 1.483~m to 0.870~m, corresponding to a relative improvement of approximately 41.3\%. This result demonstrates the benefit of incorporating measurement-noise tuning in addition to bias-instability tuning for this sequence.

\vspace{-0.5\baselineskip}

\section{Conclusion}
In this work, the covariance process noise matrices for discrete and continuous systems are formulated and analyzed. The relationship between the power spectral density and calibrated Allan variance is used to design a tuning approach for inertial-based Kalman filter. Specifically, the tuning of bias-instability parameters is mainly concerned, while the OpenVINS experiment further considers simultaneous tuning of the IMU measurement-noise parameters. The process of obtaining calibrated parameters from the Allan variance is also articulated in detail. The INS/GNSS integrated navigation system, LiDAR-inertial odometry, and visual-inertial odometry are used to show the feasibility of the proposed parameter tuning method. Future work will focus on more efficient tuning methods.

\vspace{-0.75\baselineskip}
\section*{Acknowledgements}
This work was supported by National Natural Science Foundation (62303310, U25A20477).


\bibliographystyle{unsrt}
%\bibliography{ref}
\vspace{-0.75\baselineskip}

\end{document}